\documentclass[10pt, a4paper]{article}

\usepackage[final]{lrec2026} %

\usepackage{latexsym}
\usepackage{amssymb}
\usepackage{amsmath}
\usepackage{amsthm}
\usepackage{booktabs}
\usepackage{enumitem}
\usepackage{graphicx}
\usepackage{color}
\usepackage{multirow}
\usepackage{adjustbox}
\usepackage{xcolor}

\usepackage{comment}
\usepackage{todonotes}
\usepackage{multirow}

\titlespacing*{\paragraph}{0pt}{1ex}{0pt}
\usepackage{listings}
\usepackage{xcolor}
\usepackage{hyperref}
\definecolor{myred}{RGB}{206, 50, 16}
\definecolor{mygreen}{RGB}{20, 157, 51}
\newcommand{\deltaneg}[1]{\scriptsize{\bf \textcolor{myred}{(-{#1})}}}
\newcommand{\deltapos}[1]{\scriptsize{\bf \textcolor{mygreen}{(+{#1})}}}

\definecolor{promptgray}{gray}{0.95}
\lstdefinelanguage{Prompt}{
    morecomment=[l]{\#},
}
\title{Reasoning over Object Descriptions Improves Coreference Resolution in Task-Based Dialogue Systems}

\name{Oier Ijurco, Oier Lopez de Lacalle} 

\address{HiTZ Center - Ixa, University of the Basque Country UPV/EHU \\
         \{oier.ijurco,oier.lopezdelacalle\}@ehu.eus\\}

\abstract{
Task-based dialogue systems assist users in achieving specific goals, such as executing actions or retrieving information, through natural language interactions. %
Accurate coreference resolution is essential, as it involves identifying object references within the dialogue—a task that becomes increasingly challenging in visually grounded environments characterized by complex scenes and diverse object metadata.
However, coreference resolution in task-based dialogue remains limited by poor generalization across domains and heavy reliance on supervised models that often overfit to dataset-specific artifacts. In this work, we propose a unimodal test-time reasoning approach that enables large language models (LLMs) to reason over detailed object metadata and dialogue history to improve coreference resolution. Empirical results on the SIMMC 2.1 dataset demonstrate that LLMs can generate step-by-step reasoning processes that effectively align dialogue context with objects present in the scene. Extensive experiments highlight the models' ability to link conversations and objects accurately. Moreover, we show that test-time reasoning under few-shot settings generalizes effectively to unseen scenarios and novel objects, outperforming encoder-based supervised methods in cross-domain evaluations. These findings underscore the critical role of structured metadata and careful prompt engineering in enhancing the robustness and generalization of task-oriented dialogue systems.
 \\ \newline  }

\begin{document}

\maketitleabstract

\section{Introduction}

\begin{figure}[!t]
\centering
\includegraphics[width=7.5cm]{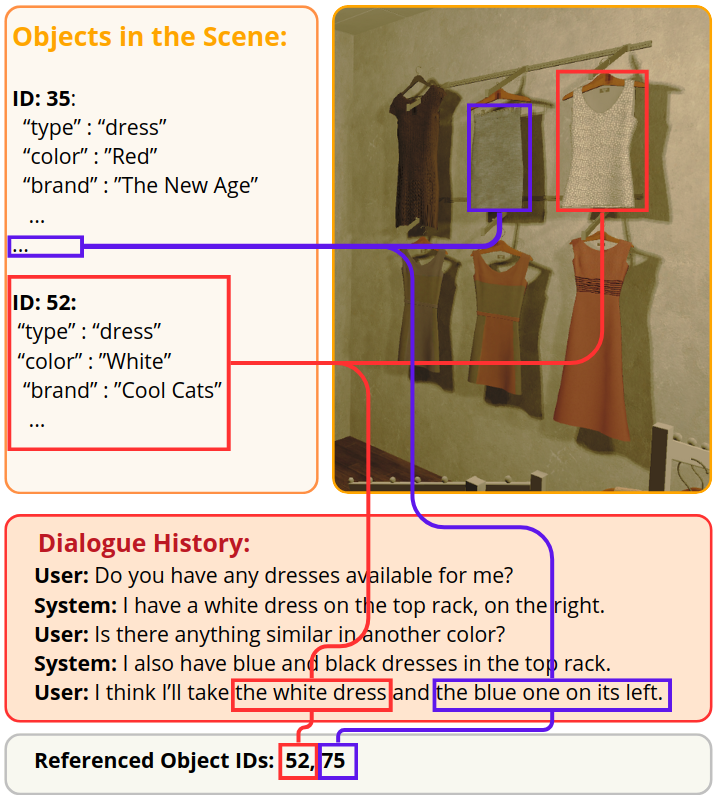}
\caption{Example of the coreference resolution task in SIMMC 2.1. Given the image of the scene, current object metadata, and dialogue history, the model must identify the object references in the last utterance of the user (e.g. \textbf{"the white dress"} with \textbf{object id 52}). }
\label{fig:introTask}
\end{figure}

Task-based dialogue systems aim to assist users in achieving specific goals, such as executing actions or retrieving information, via natural language interaction. In these settings, accurate coreference resolution and entity linking \cite{ng2002improving, lee2017end} are critical, as systems must correctly identify and resolve references to entities mentioned throughout the dialogue. This challenge is intensified in visually grounded environments, where referenced objects may appear in complex scenes and be described using diverse metadata. Resolving such references requires reasoning over both the dialogue history and multimodal context. As illustrated in Figure~\ref{fig:introTask}, effective resolution depends on a comprehensive understanding of the scene, including the description of objects in the metadata.

State-of-the-art coreference resolution methods for task-oriented dialogue systems \cite{ni2023recent} primarily rely on encoder-based architectures fine-tuned on annotated data. While effective within-domain, these models are data-intensive~\citep{huang2021uniter, lee2022tackling} and generalize poorly beyond the training distribution. Recent two-stage approaches, where an encoder processes the dialogue and scene context, followed by a reasoning decoder trained on structured outputs face similar challenges~\citep{long2023improving}. These systems often overfit to dataset-specific artifacts, such as domain identifiers or object templates, limiting their ability to generalize to unseen objects or domains. As a result, their robustness remains limited, particularly in real-world settings where labeled data is scarce.

To address the challenges of data scarcity and domain dependence, we investigate the use of generative large language models (LLMs) for coreference resolution in task-oriented dialogue, grounded in object and scene metadata. Unlike encoder-based architectures, LLMs offer greater flexibility in reasoning over diverse inputs, particularly when structured metadata is combined with natural language prompts. We hypothesize that, when appropriately prompted, LLMs can effectively integrate dialogue context with object and scene descriptions, enabling improved generalization to new domains and scenarios with limited or no annotated data.

Our study focuses on the SIMMC 2.1 dataset~\citep{kottur2023overview}, a benchmark for multimodal, task-oriented dialogue that features diverse scenes, rich object metadata, and multi-turn conversations. Although the dataset is multimodal, we adopt a unimodal (text-only) approach in this work. As illustrated in Figure~\ref{fig:introTask}, the coreference resolution task requires the model to identify object references in a user’s utterance, given the dialogue history, object metadata, and an image of the scene. We evaluate LLM performance under various prompting strategies (zero-shot, few-shot, and test-time reasoning) and show that test-time reasoning consistently improves accuracy across model variants and settings.

We further investigate cross-domain generalization, finding that fine-tuned LLMs are capable of transferring knowledge across domains. Additionally, ablation studies highlight the importance of different metadata components, and comparisons with state-of-the-art encoder-based models reveal that combining object metadata with test-time reasoning is essential for accurate reference resolution. Notably, LLMs equipped with test-time reasoning exhibit strong performance in scenarios where traditional approaches often fail.
As result of our experiments, we find that: 

\paragraph{1) Test-time reasoning enhances coreference resolution in task-oriented dialogue:}~While LLMs often struggle to incorporate rich metadata across tasks, our experiments demonstrate that test-time reasoning significantly improves performance. On the SIMMC 2.1 dataset, LLMs with reasoning prompts produce step-by-step inferences that effectively align dialogue context with object metadata.
\paragraph{2) The proposed approach shows strong cross-domain generalization and transfer capabilities:}~Cross-domain evaluations reveal that our method generalizes well to unseen objects, addressing key limitations of encoder-based baselines. Additionally, supervised fine-tuning not only enhances in-domain performance but also facilitates cross-domain transfer, with models trained in one domain showing gains in previously unseen domains.
\paragraph{3) Reformulation of structured metadata description into natural language is helpful:}~We show that the representation of object metadata has a substantial impact on model performance. Specifically, we show that natural language formulations of structured data (e.g. JSON) align better with the model, enabling more effective integration of relational and contextual information.

\section{Related Work}

\paragraph{Situated Conversational Agents}\; Situated conversational agents interact with users in environments where spatial, visual \cite{antol2015vqa, das2017visual}, and temporal reasoning \citep{thomason2020vision} is often needed. Recent advances and datasets force models to link entities mentioned in dialogue to those present in a dynamic context. Existing datasets ~\citep{das2017visual, saha2018towards,chen2019jddc,zhao2021jddc} challenge models to ground language in multimodal dynamic environments have highlighted the complexity of real-world conversational understanding~\citep{kopp2005conversational, alnefaie2021overview}.
However, prior work often uses systems which depend on extensive task-specific supervision and struggle to scale or generalize. To effectively navigate and respond in these complex settings, such agents must understand referring expressions and resolve entities across multiple modalities to maintain coherent conversations. This makes coreference resolution a central capability for such agents.

\paragraph{Coreference Resolution}\; Coreference resolution is a crucial challenge in NLP, with early approaches relying on rule-based or statistical methods \citep{soon2001machine, ng2002improving}. With the appearance of neural models, encoder-based architectures have been used to tackle the challenge \citep{lee2017end}. These models perform well on standard textual datasets but struggle in multimodal settings where visual and structured context are critical~\citep{moon2020situated}. %
Many multimodal tasks in the literature are related to Multimodal Coreference Resolution, often grouped under the term of Visual Grounding \cite{sun2022visual}, such as Referring Expression~\cite{kazemzadeh2014referitgame, qiao2020referring}, which involves locating the image region that corresponds to a given textual description. Multimodal Entity Linking further extends the challenges of Coreference Resolution into cross-modal domains, where entities must be linked across text and images \citep{adjali2020multimodal, gan2021multimodal, song2024dual, alonso2025vision}. %
Evaluating such complex models and representations requires datasets that reflect the challenges of multimodal reasoning and entity linking.

\paragraph{Task-oriented Dialogue Datasets}\; %
Datasets like MultiWOZ \citep{eric2019multiwoz}, SIMMC \citep{kottur2023overview}, and challenges like DSTC \citep{henderson2014second, williams2014dialog, williams2016dialog} have driven research in task-oriented dialogue by providing annotated multi-turn conversations. SIMMC 2.1 in particular introduces a multimodal setup,
making it a rich benchmark for evaluating multimodal coreference. These datasets highlight the limitations of static, fully trained models and expose the need for dynamic reasoning capabilities, particularly when facing out-of-distribution examples at inference. This motivates recent interest in context engineering strategies that adapt models to novel situations on the fly.

\paragraph{Context Engineering}\; 
Recent work highlights context engineering~\citep{mei2025survey} as a key paradigm for improving model robustness and adaptability without retraining. Techniques like test-time training \citep{sun2020test, akyurek2024surprising}, retrieval-augmented generation \citep{lewis2020retrieval, gao2023retrieval}, and in-context learning \citep{brown2020language, dong2024survey} use available inputs to adapt models during inference. In dialogue systems, such mechanisms are under-explored. However, they hold significant promise in enabling conversational agents to better resolve references, reason about new contexts, and personalize responses, all of which are vital for situated and multimodal interactions.

\section{Problem Formulation}

\begin{figure*}[!t]
\centering
\includegraphics[scale=0.4]{reasoning_approach3.png}
\caption{Example of our approach for coreference resolution task in SIMMC 2.1.  The LLM receives as input instruction to solve the task (\textbf{Task Instruction}) along with some reasoning steps that incorporate object descriptions (\textbf{Metadata}) and the previous dialogue and object references (\textbf{Dialogue History}). The LLM generates some reasoning that delivers the referenced object IDs by the user (\textbf{Referenced IDs}). In this case, the user is asking for information about \textit{the brown table} that is linked to \texttt{ID:6} in metadata. The LLM should return   \texttt{<SOM>6<EOM>} after reasoning about the input. }

\label{fig:approach}
\end{figure*}

\paragraph{SIMMC 2.1 Dataset}\; %
We base our study on the SIMMC 2.1 dataset~\citep{kottur2021simmc, kottur2023overview}, a task-oriented dialogue corpus designed for multimodal assistant-user interactions in realistic shopping scenarios. The dataset features complex scenes with an average of 19.7 objects arranged realistically, making it an ideal testbed for robust coreference resolution. 

The dataset comprises two distinct domains: Fashion and Furniture. While both domains share a common structure, each introduces unique objects and scenarios that do not appear in the other domain. This domain-specific variation enables a controlled evaluation of model generalization, requiring systems to adapt to novel object types and contextual cues. %

Each dialogue in the dataset is grounded in a visual environment and includes a combination of images, 3D spatial coordinates, and detailed object metadata. Dialogues are multi-turn and task-driven, encompassing user intents such as item requests, comparisons, and attribute-based inquiries. In total, the dataset comprises 11,244 dialogues and approximately 117,000 utterances.

Formally, a dialogue can be represented as a sequence of tuples: $D := {(U_t, A_t, M_t, S_t, B_t)}_{t=1}^{r}$, where $U_t$ and $A_t$ denote the user and assistant utterances at turn $t$. $M_t$ represents the multimodal context, including indices of referred objects, and $S_t$ encodes the scene context, such as object attributes and spatial metadata (see \S\ref{sec:approach} for details). The belief state $B_t$ includes dialogue acts and slot-value pairs, but is not utilized in this work.

Although the benchmark defines four tasks, in this work, we focus on the coreference resolution task, which is detailed in the following section.

\paragraph{Coreference Resolution Task}\;

In SIMMC, the task requires identifying which objects, if any, are referenced in the user's most recent utterance within a multi-turn dialogue. Each dialogue is grounded in a scenario with a scene \( S \), containing candidate objects \( \mathcal{O} = \{o_1, o_2, \ldots, o_n\} \), each associated with structured metadata \( \text{meta}(o_i) \), a dialogue history \( H = \{u_1, s_1, \ldots, u_{t-1}, s_{t-1}\} \), where \( u_i \) and \( s_i \) are the user and system utterances at turn \( i \), and the current user utterance \( u_t \). The task is to predict, for each object \( o_i \in \mathcal{O} \), a label \( y_i \in \{0, 1\} \) indicating whether \( o_i \) is referred to in \( u_t \) given \( H \), \( S \) and \( \text{meta}(o_i) \). If no object is referenced, then \( y_i = 0 \) for all \( i \). This frames the coreference resolution task as a per-object binary classification problem within each scene. Accurate resolution requires joint reasoning over the dialogue context, multimodal scene information, and fine-grained object metadata.

\section{Proposed Approach}
\label{sec:approach}
We adopt a generative large language model (LLM) based approach that reframes the problem of coreference resolution problem as a reference identification task. Contrary to the previous approaches, in which they focus on fine-tuning models on annotated data, we adopt in-context learning and chain-of-thought approaches to fully harness the reasoning capabilities of current LLMs. Our approach allows us to generalize the coreference resolution task to unseen objects and scenes, which is exemplified in Figure~\ref{fig:approach}, and proceeds as follows. 
At each dialogue turn, the LLM is prompted with information that includes%
: 1) A complete list of objects in the scene, each with detailed \textbf{metadata} such as the \texttt{ID}, \texttt{type}, \texttt{color}, \texttt{brand}, and  \texttt{location}, among others.  2) The \textbf{dialogue history} between the user and the system. The dialogue history comes with all the object IDs mentioned earlier in the conversation\footnote{Previously mentioned objects are coded as \texttt{<SOM> objectID <EOM>}.}. Each turn is finished with the last utterance of the user. 
3) Finally, \textbf{task-specific instructions} outlining the required actions for the model to complete the task are given. In addition, a structured set of reasoning steps is provided to aid the model in aligning contextual information appearing in the dialogue and the detailed description of objects occurring in the scene. Further details of the reasoning steps are given in the follow-up section (\S\ref{sec:reasoning}).

The LLM is then prompted to output the object ID(s) mentioned in the last user utterance. In case the user does not mention any object, the model is instructed to output an empty sets of objects (i.e., \texttt{<SOM><EOM>})\footnote{E.g., "\texttt{User:} \emph{Hi, do you have any jackets today?}" would be an utterance without mentioning any object, in which empty set of object IDs should be returned.}. The response is also considered as type of prediction over the set of possible object IDs.

The proposed formulation allows the model to reason directly over the metadata and conversational context without any fine-tuning on annotated data. We experimented with various prompting strategies, including few-shot examples and chain-of-thought reasoning to enhance the performance of the model. The approach is flexible, domain-agnostic, and well-suited to scenarios where traditional models struggle due to limited supervision or unseen objects in a new domain.

\subsection{Reasoning steps}
\label{sec:reasoning}

Alignment of the conversational context and information contained in the metadata is not trivial for the current LLMs, but it is required to correctly reference the mentioned objects. Thus, the aim of this additional prompt is to generate a concrete inferring process step-by-step and implement corresponding actions on the conversational history and metadata to get information about object IDs. As Figure~\ref{fig:approach} shows, the LLM takes the task instructions, the reasoning steps, and generates the reasoning that aligns dialogue history and metadata to provide the referenced object ID. The LLM needs to go through the subsequent step-by-step deliberation process of 5 steps: 
\begin{enumerate}
    \item \textbf{Identification of referential expressions}: In this step, the model extracts the noun phrases, pronouns and any descriptive phrases in the dialogue that may mention any object of the scene.

    \item \textbf{Getting object attributes for the referential expressions}: In this step, the goal is to gather relevant metadata (e.g., type, color, location) for each identified object mention.

    \item \textbf{Mapping the current object IDs to object attributes}: Once the relevant object attributes are obtained, the model attempts to match object IDs to the extracted attributes, selecting only those with strong alignment.

    \item \textbf{Ambiguity resolution}: In cases where multiple objects meet the referential criteria, the model incorporates dialogue history to resolve ambiguity and narrow down the selection.

    \item \textbf{Output the identified object IDs}: Finally, the model returns confident object IDs between \texttt{<SOM>} and \texttt{<EOM>} tags. Otherwise, it outputs nothing to avoid errors.
\end{enumerate}

These reasoning steps reflect the way a human would reason to perform this same task and predict the objects that are being referenced in the given utterance in the dialog.

\section{Experimental Setting}
\label{sec:exp_setting}
We conducted our experiments using a variety of strong open models that are publicly available. The evaluated models include the Llama 3 family~\cite{grattafiori2024llama3herdmodels}, which we tested Llama 3.1-8B and 3.3-70B models, instructed Gemma-7B \cite{team2024gemma}, instructed Mistral-7B~\cite{jiang2023mistral7b}, instructed Qwen 2.5-7B~\cite{qwen2025qwen25technicalreport}, Qwen3-4B~\cite{yang2025qwen3} and the distilled version of DeepSeek-R1 to Llama3.1-8B~\cite{deepseekai2025deepseekr1incentivizingreasoningcapability}.

All open-weight models were evaluated in a computing cluster using four NVIDIA A100 80GB GPUs. The entire set of experiments was estimated to have required approximately 750 GPU hours.

We conducted two main sets of experiments to evaluate the effectiveness of the large language models for coreference resolution in task-oriented dialogue settings. The first set of experiments is devoted to assessing the contribution of test-time reasoning to understanding object descriptions in the metadata and the dialogue history (\S\ref{sec:reasoning_results}). We explore different prompting strategies to measure the effectiveness of test-time reasoning. In addition, to ensure the models leverage object descriptions effectively, we conduct an ablation study on the availability of information sources, such as previous object references in the dialogue and access to the object's metadata (\S\ref{sec:information}).

In the second set of experiments, we measure the ability of the LLMs to generalize across domains and handle unseen objects (\S\ref{sec:xdomain_results}). To this end, we perform cross-domain evaluations, in which models are trained in one domain (say furniture domain) and evaluated in the other domain (say fashion). This setting ensures no seen objects occur in the testing domain. 
In addition, we compare the generative models against state-of-the-art encoder models under the same conditions to measure their robustness and adaptability to new, unseen domains. All experiments were performed using the SIMMC 2.1 dataset, focusing on both the Fashion and Furniture domains. 

To efficiently fine-tune the model in the second set of experiments, we employ parameter-efficient fine-tuning (PEFT) \cite{han2024parameter} using LoRA (Low-Rank Adaptation) \cite{hu2022lora}. Specifically, we fine-tune the Qwen2.5-7B-Instruct and the Llama3.3-70B-Instruct models with LoRA configurations: rank (r) = 4, LoRA alpha = 8, and dropout = 0.05. %

The dataset includes approximately 11K task-oriented dialogues (around 117K utterances) between an assistant and a user, situated in commercial store environments enriched with scene images and item metadata. The dataset spans the fashion (7.2K dialogues) and furniture (4K dialogues) domains, with the fashion domain presenting more complexity due to higher item density and visual overlap in scenes. On average, dialogues contain 10.4 utterances, with assistant turns averaging 13.7 words. Each dialogue mentions about 4.7 objects, while scenes contain an average of 19.7 objects. The average number of objects per furniture scene is slightly over 10, whereas in the fashion domain the mean exceeds 30, reaching over 55 items in some cases. The data is split into Train (64\%, 7307 dialogues), Dev (5\%, 563), Dev-test (15\%, 1687), and Std-test (15\%, 1287). We use the Dev-test set for all our evaluations, which comprises 8609 instances.

In all the experiments, the few-shot examples have been selected from the training split to reflect a diverse range of scenarios. This ensures that the model is exposed to various challenges it may encounter during inference. Specifically, the examples include: one instance where no objects are referenced, another where the currently referred objects have been previously mentioned in the dialogue, and a final case where the current objects are being referred to for the first time. The examples span across both the Fashion and Furniture domains, further contributing to a more representative and balanced demonstration of the model's capabilities.

Additionally, we conduct a qualitative analysis to better understand the factors influencing model behavior (\S\ref{sec:information}). Specifically, we examine how input types, metadata formats, and prompting strategies affect performance and interpretability. These analyses include evaluating information access in resolving references, comparing metadata formats and assessing prompt design effects on predicted object references. Together, these studies offer deeper insights into the interpretability of LLM behavior in task-oriented dialogue systems.

\section{Results}
\label{sec:main_results}

\subsection{The Impact of Reasoning}
\label{sec:reasoning_results}

In the subsequent experiments, we evaluate the effectiveness of test-time reasoning by comparing our approach against different prompt completion strategies. We consider three types of prompts: \textbf{Zero-shot}, where the LLM receives only the task instructions along with dialogue history and object metadata, but no input-output examples; \textbf{Few-shot}, which extends the zero-shot setting by including three coreference resolution examples from both the fashion and furniture domains; and \textbf{Few-shot+Reasoning}, in which these few-shot examples are augmented with an additional reasoning step that explicitly explains how object metadata relates to the dialogue context. The full prompt structure and a complete few-shot instance used in our test-time reasoning approach are provided in Appendix~\ref{appendix:instructions}.

\begin{table}[t]
\centering
\begin{adjustbox}{max width=\columnwidth}
\begin{tabular}{p{0.05cm}lp{0.5cm}ccc}
\toprule
& \textbf{Model} & \textbf{Size} &\textbf{Precision} & \textbf{Recall} & \textbf{F1 Score} \\
\midrule
& Random & & 2.68 & 49.87 & 5.09 \\
\midrule
\multirow{7}{*}{\rotatebox{90}{Zero-shot}}
& Qwen3 & 4B & \textbf{29.01} & 63.91 & 39.90 \\
& Gemma & 7B & 16.93 & 52.74 & 25.64 \\
& Mistral & 7B & 25.70 & 66.62 & 37.09 \\
& Qwen2.5 & 7B & 24.14 & 71.66 & 36.12 \\
& Llama3.1 & 8B & 27.74 & 63.03 & 38.53 \\
& DS-R1 & 8B & 22.98 & 66.77 & 34.19 \\
& Llama3.3 & 70B & 26.61 & \textbf{80.10} & \textbf{39.95} \\
\midrule
\multirow{7}{*}{\rotatebox{90}{Few-shot}}
& Qwen3 & 4B & 39.13\deltapos{10.12} & 59.15\deltaneg{4.76} & 47.10\deltapos{7.2} \\
& Gemma & 7B & 19.58\deltapos{2.65} & 41.38\deltaneg{11.36} & 26.58\deltapos{0.94}\\
& Mistral & 7B & 29.14\deltapos{3.44} & 62.65\deltaneg{3.97} & 39.77\deltapos{2.68} \\
& Qwen2.5 & 7B & 28.84\deltapos{4.7} & 70.05\deltaneg{1.61} & 40.86\deltapos{4.74} \\
& Llama3.1 & 8B & 30.34\deltapos{2.6} & 68.39\deltapos{5.36} & 42.03\deltapos{3.5} \\
& DS-R1 & 8B & 25.31\deltapos{2.33} & 67.17\deltapos{0.4} & 36.77\deltapos{2.58} \\
& Llama3.3 & 70B & \textbf{39.32}\deltapos{12.71} & \textbf{73.12}\deltaneg{6.98} & \textbf{51.14}\deltapos{14.34} \\
\midrule
\multirow{7}{*}{\rotatebox{90}{Reasoning}}
& Qwen3 & 4B & \textbf{59.98}\deltapos{20.85} & 45.21\deltaneg{13.94} & 51.55\deltapos{4.45} \\
& Gemma & 7B & 37.86\deltapos{18.28} & 43.24\deltapos{1.86} & 40.37\deltapos{13.79} \\
& Mistral & 7B & 39.82\deltapos{10.68} & 52.93\deltaneg{9.72} & 45.45\deltapos{5.68} \\
& Qwen2.5 & 7B & 49.90\deltapos{21.06} & 57.32\deltaneg{12.73} & 53.35\deltapos{12.49} \\
& Llama3.1 & 8B & 52.16\deltapos{21.82} & 59.23\deltaneg{9.16} & 55.47\deltapos{13.44} \\
& DS-R1 & 8B & 53.50\deltapos{28.19} & 43.92\deltaneg{23.25} & 48.24\deltapos{11.47} \\ 
& Llama3.3 & 70B & 58.73\deltapos{19.41} & \textbf{61.37}\deltaneg{11.75} & \textbf{60.02}\deltapos{8.88} \\
\bottomrule
\end{tabular}
\end{adjustbox}
\caption{Results of the models evaluated in different prompt settings (zero-shot, few-shot, and test-time reasoning). Numbers in parentheses show the differences with the previous prompt technique: zero-shot vs few-shot, and few-shot vs reasoning. DS-R1 refers to DeepSeek-R1-Distill-Llama-8B}
\label{tab:model_metrics}
\end{table}

Table~\ref{tab:model_metrics} reports the precision, recall, and F1 scores obtained in SIMMC 2.1 for several models evaluated across the different prompt settings. 

\paragraph{Few-shot improvement}\; In terms of the F1 score, the results demonstrate a consistent improvement when examples are incorporated into the prompt. Note that only 3 examples are added in the prompt. Overall, all models with 7B to 8B parameters exhibit an improvement between 1 and 4.5 points compared to zero-shot versions. The 70B Llama model shows an even greater enhancement, achieving an increase of more than 14 points. 
The performance of Qwen3 is particularly noteworthy: despite being significantly smaller in size, it achieves a comparable F1 score to the 70B Llama model in the zero-shot setting. Moreover, Qwen3 exhibits a significant improvement of 7.2 points when few-shot examples are incorporated into the prompt.
Analysis indicates that the improvement in F1 is primarily driven by an increase in precision, at the cost of recall. This suggests that, under the few-shot setting, the models tend to respond to fewer IDs and have begun to better infer how to perform the task.

\paragraph{Few-shot vs Reasoning}\; The application of reasoning on top of few-shot learning also yields positive results in terms of F1 scores. All evaluated models consistently demonstrate significant improvements, with the observed gains being even more pronounced than the differences between zero-shot and few-shot settings. Across all models, the average improvement is approximately 10 points. Notably, the Qwen2.5 and Llama3.1 models exhibit particularly strong improvements, achieving performance levels close to that of the 70B Llama3.3 model. In terms of precision, the improvements are substantial, with the models achieving an average increase of 19 points. However, this precision gain comes at the cost of Recall, as reasoning trends to reduce overprediction of objects (\S\ref{sec:numObj}).

\paragraph{Model size}\; Although a detailed analysis has not been conducted, the experiments suggest that model size plays an important role. As the model size increases, the ability to correctly perform reasoning steps also improves. Despite being specifically designed for reasoning, Deepseek-R1 does not achieve strong results.

\subsection{Cross-domain Experiments}
\label{sec:xdomain_results}

The following experiments investigate the cross-domain generalization capabilities of LLMs, with a particular emphasis on their performance when exposed to novel scenes and entities absent during training. Specifically, we conduct two evaluations: in the first, the model is trained solely on data from the fashion domain and tested on the furniture domain (\textbf{Fashion $\rightarrow$ Furniture}); in the second, the training and testing domains are reversed, with the model trained on furniture data and evaluated on fashion (\textbf{Furniture $\rightarrow$ Fashion}). In both cases, the test sets contain previously unseen objects, enabling a robust assessment of the model’s generalization across domains. We selected Qwen2.5 and Llama3.3 models for our experiments. We compare the generative models against a strong encoder-based model (BART$_{\text{encoder}}$) under the same conditions to measure robustness.
The results of the experiments are shown in Table~\ref{tab:combined-results}a and Table~\ref{tab:combined-results}b.

\begin{table}[h]
\centering
\begin{small}
\begin{tabular}{llcc}
\toprule
\multicolumn{4}{c}{\textbf{(a) Trained on Fashion}} \\
\midrule
\textbf{Setting} & \textbf{Model Name} & \textbf{Fashion} & \textbf{Furniture} \\
\midrule
& BART$_{\text{encoder}}$ & 70.93 & 6.41 \\
\midrule
\multirow{4}{*}{FS} & Qwen2.5  & 39.86 & 49.85 \\
                    & Llama3.3 & 52.05 & 55.54 \\
                    & Qwen2.5$_{\text{+R}}$ & 52.60 & 63.22 \\
                    & Llama3.3$_{\text{+R}}$ & 59.96 & 59.28 \\
\midrule  
\multirow{4}{*}{SFT} & Qwen2.5 & 55.98 & 67.06 \\
                     & Llama3.3 & \textbf{73.62} & \textbf{75.51} \\
                     & Qwen2.5$_{\text{+R}}$ & 55.04 & 64.92 \\
                     & Llama3.3$_{\text{+R}}$ & 66.06 & 65.04 \\
\bottomrule
\toprule

\multicolumn{4}{c}{\textbf{(b) Trained on Furniture}} \\
\midrule
\textbf{Setting} & \textbf{Model Name} & \textbf{Fashion} & \textbf{Furniture} \\
\midrule
& BART$_{\text{encoder}}$ & 5.44 & \textbf{78.78} \\
\midrule
\multirow{4}{*}{FS} & Qwen2.5  & 35.27 & 52.63 \\
                    & Llama3.3 & 46.46 & 56.09 \\
                    & Qwen2.5$_{\text{+R}}$ & 45.18 & 60.59 \\
                    & Llama3.3$_{\text{+R}}$ & 54.80 & 60.85 \\
\midrule  
\multirow{4}{*}{SFT} & Qwen2.5  & 51.34 & 63.05 \\
                     & Llama3.3 & 59.66 & 64.60 \\
                     & Qwen2.5$_{\text{+R}}$ & 49.38 & 60.47 \\
                     & Llama3.3$_{\text{+R}}$ & \textbf{62.55} & 60.44 \\
\bottomrule
\end{tabular}
\end{small}
\caption{Results of models trained on (a) the \textbf{Fashion} domain and (b) the \textbf{Furniture} domain. Each is evaluated both in-domain and cross-domain. \textbf{FS} refers to the few-shot setting (no actual training). \textbf{SFT} refers to supervised fine-tuning. The subscript \textbf{+R} indicates test-time reasoning is applied.}
\label{tab:combined-results}
\end{table}

Tables are divided into prompt variants of few-shot (FS) and supervised fine-tuning (SFT). In both cases, reasoning is applied at inference time too (results of models with subscript+R).  For fine-tuning, models are provided with task instructions, metadata, and dialogue history, and they learn to generate mentioned object identifiers. 

Overall, both domains exhibit similar behaviors of models. Due to the fewer objects to predict in the Furniture domain, all models tend to achieve higher performance in this context. The results indicate that supervised classifiers based on encoder architectures (BART$_\text{encoder}$) cannot generalize effectively to out-of-domain scenarios. While these models obtain strong results in in-domain evaluations (achieving F1 scores of approximately 71 and 79 in the Fashion and Furniture domains, respectively), their performance degrades substantially when applied to unseen domains.
In contrast, models based on few-shot learning and reasoning (as shown in the FS rows) demonstrate stronger cross-domain robustness. 

The few-shot results reveal a clear imbalance between the two transfer directions. When models are given examples from the Fashion domain, they achieve higher scores when evaluated on Fashion than the opposite direction. This suggests that the Fashion domain, being more complex, provides richer examples. As a result, models exposed to Fashion few-shot examples learn more generalizable patterns that transfer effectively to the simpler Furniture domain. On the other hand, few-shot learning on Furniture lacks sufficient representational variety to generalize back to Fashion, leading to lower cross-domain scores. These findings highlight that domain complexity plays an important role in enabling LLMs to build transferable understanding from few-shot examples.

Furthermore, fine-tuned LLMs (SFT-Qwen and SFT-Llama) yield significant improvements in both in-domain and cross-domain performance over their few-shot counterparts.  The Llama models, in particular, achieve the best overall results, reaching F1 scores above 73 and 75 when trained on Fashion and tested on Fashion and Furniture, respectively (Table~\ref{tab:combined-results}a). Improvements are also observed when training on Furniture and testing on Fashion, where SFT-Llama rises from 49.78 to 59.66. Similar trends are observed for Qwen models, though with slightly lower absolute performance. It is worth noting that inference-time reasoning does not yield additional benefits for fine-tuned models, and in some cases slightly decreases performance (SFT-Llama${\text{+R}}$ and SFT-Qwen${\text{+R}}$).

\section{Analysis}

\subsection{Effect of Information Access}
\label{sec:information}

We examine the role of object descriptions in enhancing the ability of LLMs to resolve referential expressions. To assess the contribution of different information sources, we conduct a controlled ablation study across the different prompt settings. Specifically, we evaluate model performance under three input configurations: in the \textbf{All information} setting, the model receives full access to both the descriptions of all objects in the scene and the complete dialogue history; in the \textbf{No Metadata} setting, object metadata is withheld, so the model relies solely on the dialogue history to resolve references; and in the \textbf{No Object References} setting, mentions of objects within the dialogue history are removed, limiting the model's input to metadata and non-referential dialogue content. Example prompts for each configuration are shown in Appendix~\ref{appendix:ablation}.

\begin{table}[ht]
\centering
\begin{adjustbox}{max width=\columnwidth}
\begin{tabular}{llccc}
\toprule
\textbf{Setting} & \textbf{Metadata} & \textbf{Precision} & \textbf{Recall} & \textbf{F1 Score} \\
\midrule
\multirow{3}{*}{Zero-shot} 
    & All Info & 24.14 & 71.66 & 36.12 \\
    & No Metadata & 22.18 & 52.96 & 31.27 \\
    & No Objects & 16.75 & 60.82 & 26.26 \\
\midrule
\multirow{3}{*}{Few-shot} 
    & All Info. & 28.84 & 70.54 & 40.86 \\
    & No Metadata & 36.56 & 51.98 & 42.93 \\
    & No Objects & 20.66 & 56.93 & 30.31 \\
\midrule
\multirow{1}{*}{FS+Reasoning} 
    & All Info & 49.90 & 57.32 & 53.35 \\
\bottomrule
\end{tabular}
\end{adjustbox}
\caption{Qwen2.5-7B performance across different metadata representations and settings.}
\label{tab:model_metrics_qwen}
\end{table}

\begin{table}[ht]
\centering
\begin{adjustbox}{max width=\columnwidth}
\begin{tabular}{llccc}
\toprule
\textbf{Setting} & \textbf{Metadata} & \textbf{Precision} & \textbf{Recall} & \textbf{F1 Score} \\
\midrule
\multirow{3}{*}{Zero-shot} 
    & All Info & 26.61 & 80.10 & 39.95 \\
    & No Metadata & 27.94 & 55.35 & 37.13 \\
    & No Objects & 23.64 & 72.49 & 35.66 \\
\midrule
\multirow{3}{*}{Few-shot} 
    & All Info & 39.32 & 73.12 & 51.14 \\
    & No Metadata & 56.87 & 45.43 & 50.51 \\
    & No Objects & 34.69 & 70.70 & 46.54 \\
\midrule
\multirow{1}{*}{FS+Reasoning} 
    & All Info & 58.73 & 61.37 & 60.02 \\
\bottomrule
\end{tabular}
\end{adjustbox}
\caption{Llama3.3-70B performance across different metadata representations and settings.}
\label{tab:model_metrics_llama}
\end{table}

Our ablation analysis underscores the critical role of both object metadata and previously mentioned objects within the dialogue. The results in Tables~\ref{tab:model_metrics_qwen} and \ref{tab:model_metrics_llama} suggest that LLMs mainly rely on previously mentioned objects to resolve references effectively. In the zero-shot setting, removing these references reduces Qwen’s F1 score by about 10 points, while the larger Llama 70B model, although more robust, still shows a notable drop. This indicates that even larger models remain sensitive to referential context.

A comparable pattern appears in the few-shot setting: removing object references lowers Qwen’s score by around 11 points and Llama’s by about 5, confirming that reference tracking strongly influences performance regardless of model scale or supervision.

When object metadata is excluded, performance differences narrow, suggesting that models may not effectively integrate structured metadata with dialogue context. For instance, in the few-shot setting, Llama’s performance remains stable and Qwen’s slightly improves, implying that metadata can sometimes introduce noise without explicit prompt guidance.

\subsection{Metadata Representation}
\label{sec:metadata}

\begin{figure}[h]
    \centering
    \includegraphics[width=0.45\textwidth]{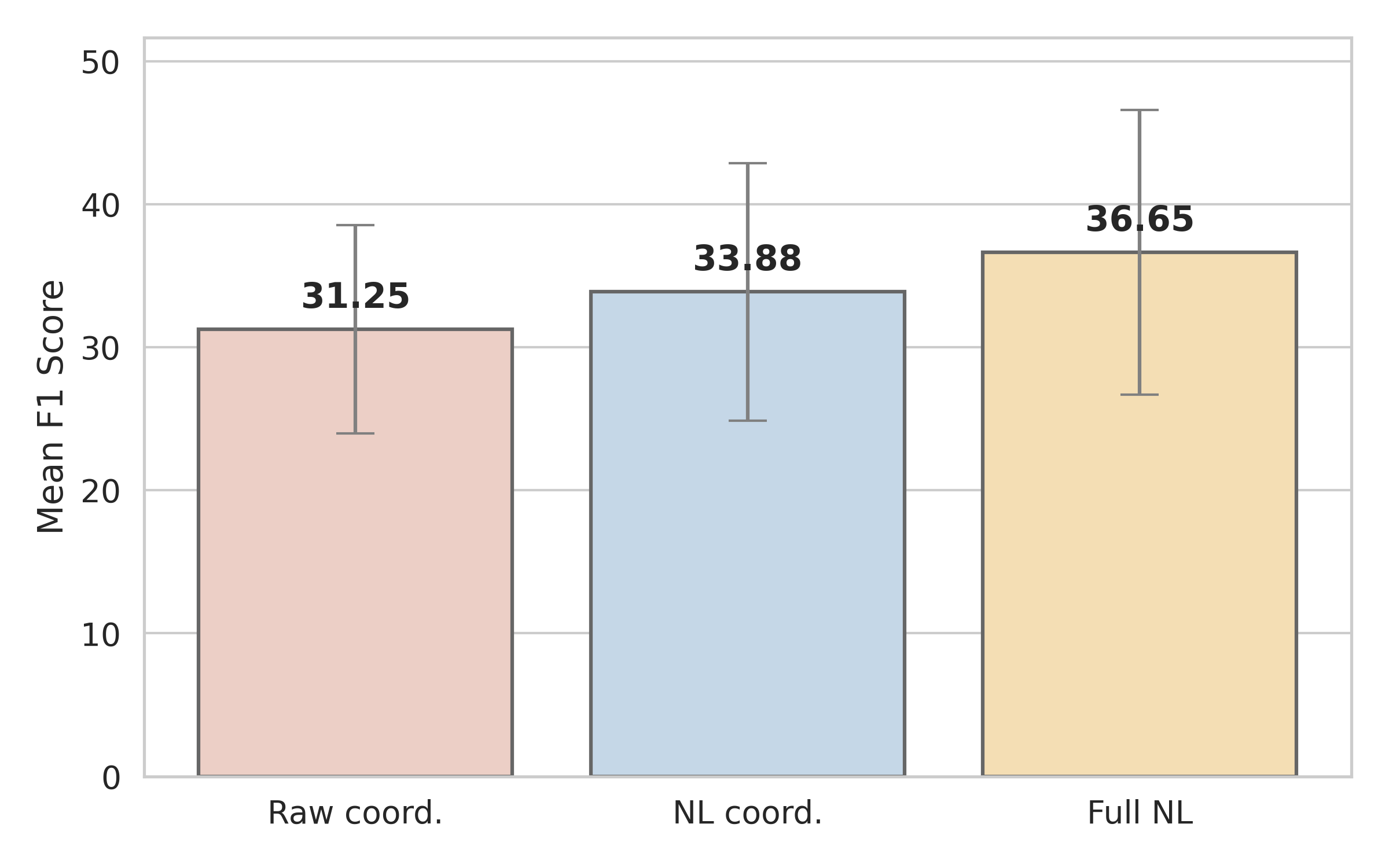}
    \caption{F1 mean scores and standard deviation of all models across different object representation types. Raw coordinates use normalized values for coordinates; NL coordinates map these values to spatial descriptors; Full NL expresses all metadata in fluent natural language.}
    \label{fig:rep}
\end{figure}

An insight from our experiments is that the way object metadata is represented substantially affects model performance. As shown in Figure~\ref{fig:rep}, expressing metadata in natural language outperforms structured formats such as JSON. While JSON provides a clear schema, it can hinder cross-attribute reasoning due to tokenization inefficiencies and limited contextual integration.

In contrast, natural language formulations align better with the model’s pretraining, enabling more effective integration of relational and contextual information. We compare three variants of metadata representations: in the first, \textbf{Raw coordinates} are encoded as normalized numerical values, an exact format that, however, forces the model to infer spatial semantics from raw numbers. The second representation, \textbf{Coordinates as natural language}, converts numerical values into coarse spatial descriptors such as "bottom-left" or "center", introducing semantically meaningful spatial cues about object location. The third variant, \textbf{Full natural language descriptions}, reformulates all metadata fields into fluent, human-readable sentences, removing key-value structures and providing the model with a more natural input format. Concrete examples of all three representation types are given in Appendix~\ref{appendix:objectRep}.

As illustrated in Figure~\ref{fig:rep}, the full natural language representation consistently outperforms the other formats, suggesting that LLMs are more adept at reasoning over unstructured, linguistically rich inputs than over structured data.

\subsection{Number of Returned Objects}
\label{sec:numObj}
\begin{figure}[h]
    \centering
    \includegraphics[width=0.45\textwidth]{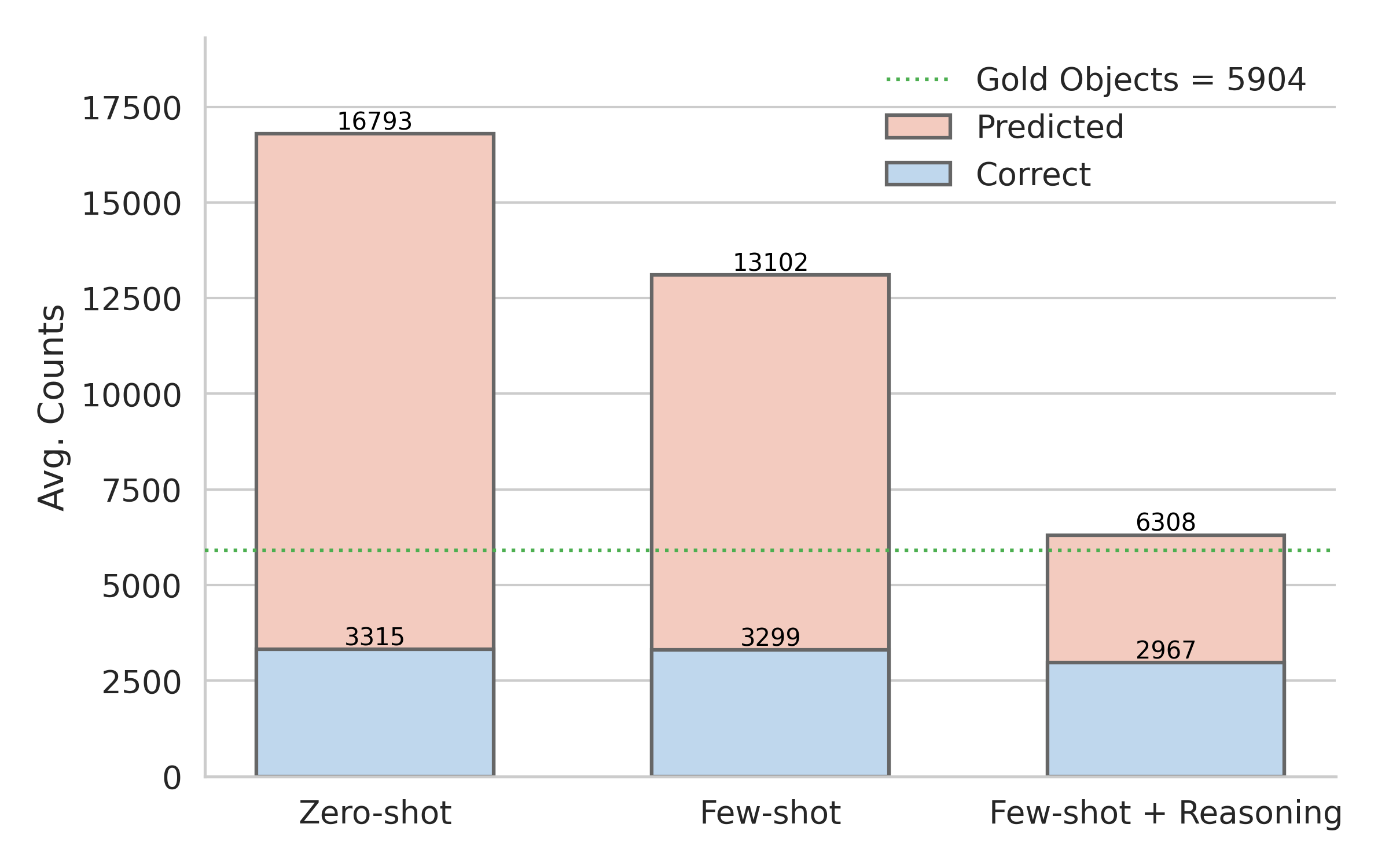}
     \caption{The average count of predicted objects and correct predictions per setting. The average is computed across all the models and settings used in the experiments.}
    \label{fig:number_objects}
\end{figure}

Figure~\ref{fig:number_objects} presents the impact of different prompting strategies on the number of predicted object references. We observe that prompts incorporating few-shot examples and explicit reasoning steps lead to fewer object IDs being predicted per utterance. This reduction is associated with increased precision and a slight decrease in recall, suggesting that the model becomes more conservative in its predictions. Rather than over-generating references, the model appears to better recognize uncertainty and abstains from producing potentially incorrect outputs. This behavior indicates a stronger alignment with task semantics, favoring precision over recall in ambiguous contexts.

\section{Conclusions}
\label{sec:conclusion}

We present a prompting-based approach for resolving object references in task-oriented dialogue by leveraging rich object metadata, structured scene context, and carefully designed prompting strategies. Our experiments demonstrate that large language models can generate effective step-by-step reasoning to align dialogue history with objects in the environment. We show that test-time reasoning combined with few-shot examples significantly improves coreference resolution performance. Cross-domain evaluations further show that our method generalizes well to unseen objects, addressing key limitations of encoder-based baselines. Additionally, supervised fine-tuning not only enhances in-domain performance but also facilitates cross-domain transfer, with models trained in one domain showing gains in previously unseen domains. Our analysis highlights the importance of representing object metadata in natural language, which better aligns with model pretraining and supports more effective integration of relational and contextual cues.

Future work may explore fine-tuning strategies that explicitly target reasoning capabilities, as well as methods for prompt compression and dynamic context selection. Integrating visual grounding modules also presents a promising way to improve performance in interactive, multimodal settings.

\section{Limitations}
\label{sec:limitations}

While our approach shows substantial improvements in coreference resolution by leveraging large language models and structured metadata, several limitations remain, highlighting opportunities for future work. 

First, our experiments are conducted exclusively on the SIMMC 2.1 dataset. Although SIMMC 2.1 provides a rich multimodal setting, the lack of comparable publicly available datasets restricts the opportunity to evaluate our findings to broader task-oriented dialogue scenarios.

Second, despite SIMMC 2.1’s multimodal nature, our method focuses solely on textual and structured metadata, omitting direct use of visual input. Incorporating visual signals into the reasoning process remains an important direction for future research.

Finally, while our results with supervised fine-tuning suggest promising cross-domain transfer capabilities, further investigation is needed to better understand the underlying mechanisms and to improve fine-tuning strategies for robustness and scalability.

\section*{Acknowledgments}

This work was supported by the European Research Council under Horizon Europe, grant number 10113572, related to LUMINOUS project, and the Spanish Ministry of Science and Innovation (AI4I/MOLVI project PID2024-157855OB-C32 and HumanAIze project AIA2025-163322-C61) funded by MICIU/AEI/10.13039/501100011033 and by ERDF, EU.

\section{Bibliographical References}\label{sec:reference}

\bibliographystyle{lrec2026-natbib}
\bibliography{lrec2026-example}

\bibliographystylelanguageresource{lrec2026-natbib}

\newpage
\appendix

\section{Prompt used for our test-time reasoning approach}
\label{appendix:instructions}
\begin{figure*}[p]
\centering
\includegraphics[width=\linewidth]{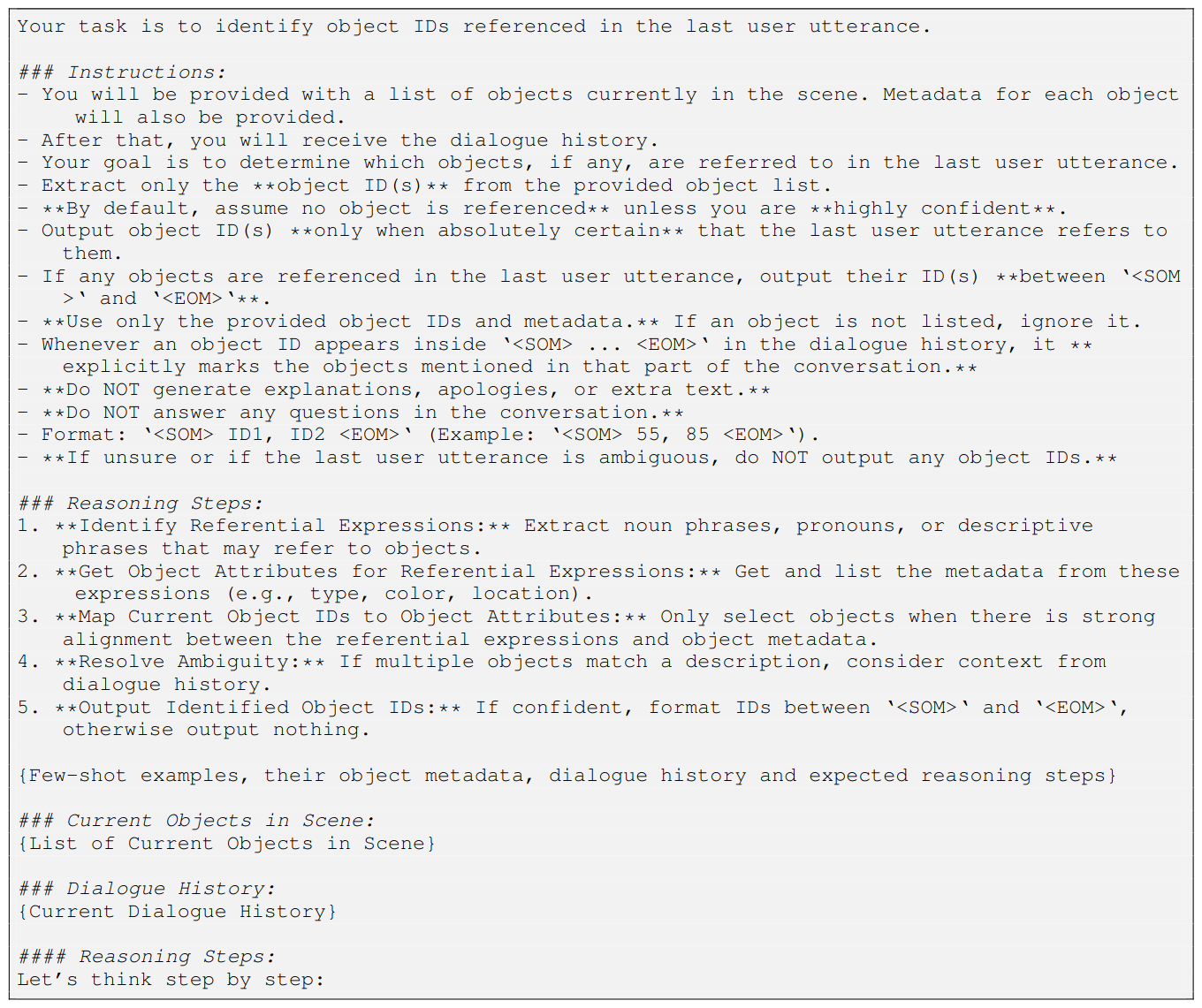}
\caption{Full prompt used for our test-time reasoning approach. The prompt is structured into three sections: task instructions and output format constraints, a chain-of-thought reasoning procedure, and few-shot examples.}
\label{fig:instructionAppendix}
\end{figure*}

\begin{figure*}[p]
\centering
\includegraphics[width=\linewidth]{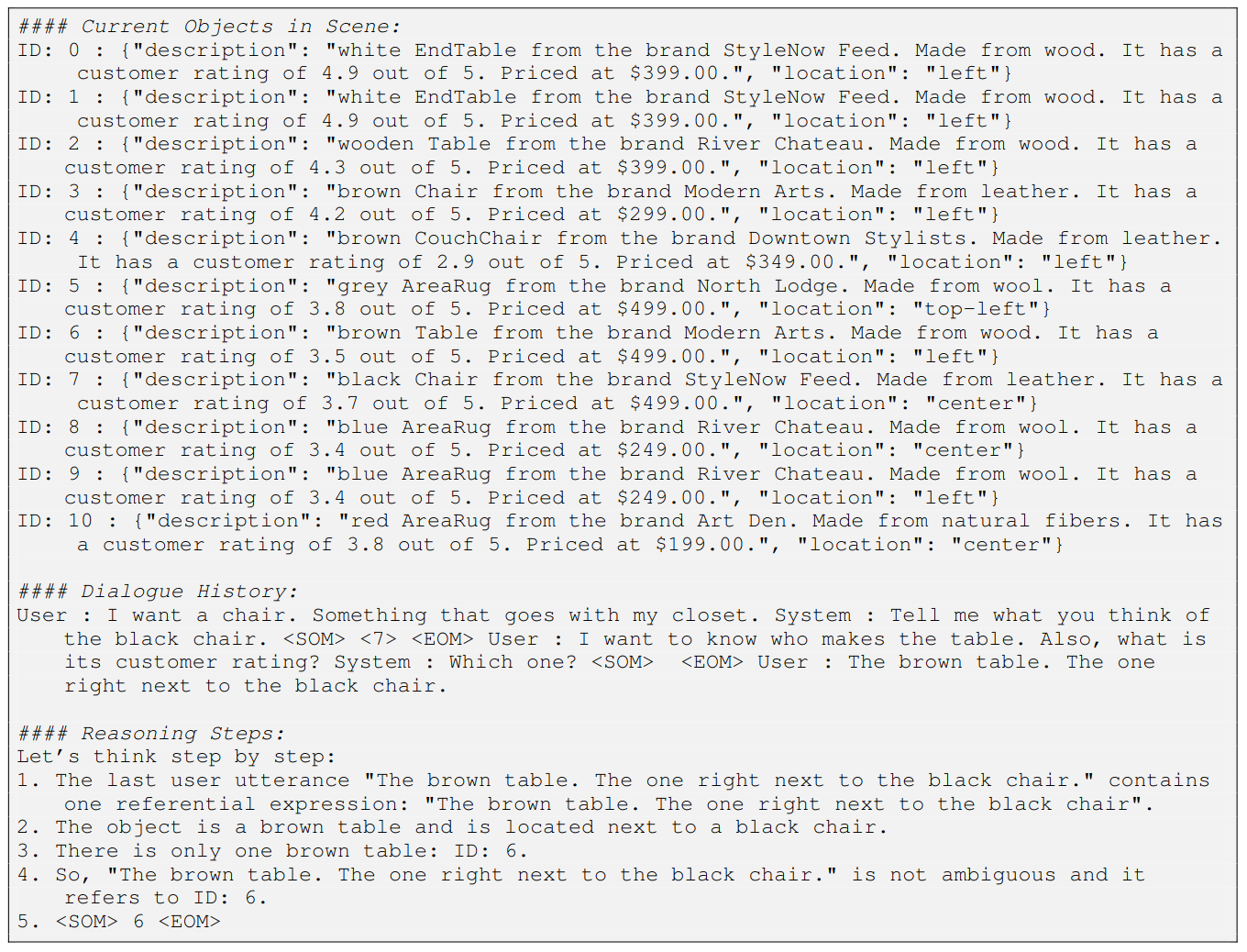}
\caption{Illustration of a complete few-shot instance as provided in the prompt, including the user utterance, object metadata, and the expected structured output. Object metadata is represented as Full Natural Language Descriptions.}
\label{fig:exampleAppendix}
\end{figure*}

Figure~\ref{fig:instructionAppendix} shows the prompt used in our test-time reasoning approach, which is structured to guide a language model through a multi-step inference process aimed at grounding user utterances in object metadata. It is divided into clearly demarcated sections: Instructions, reasoning steps, and few-shot examples. The instructions define the task, constraints, and expected output format. The reasoning steps section outlines a step-by-step procedure designed to encourage chain-of-thought reasoning, including referential expression identification, attribute mapping, and ambiguity resolution. An example of a few-shot instance can be seen in Figure~\ref{fig:exampleAppendix}.

\section{Ablation Study: Input Configurations}
\label{appendix:ablation}
\begin{figure*}[p]
\centering
\includegraphics[width=\linewidth]{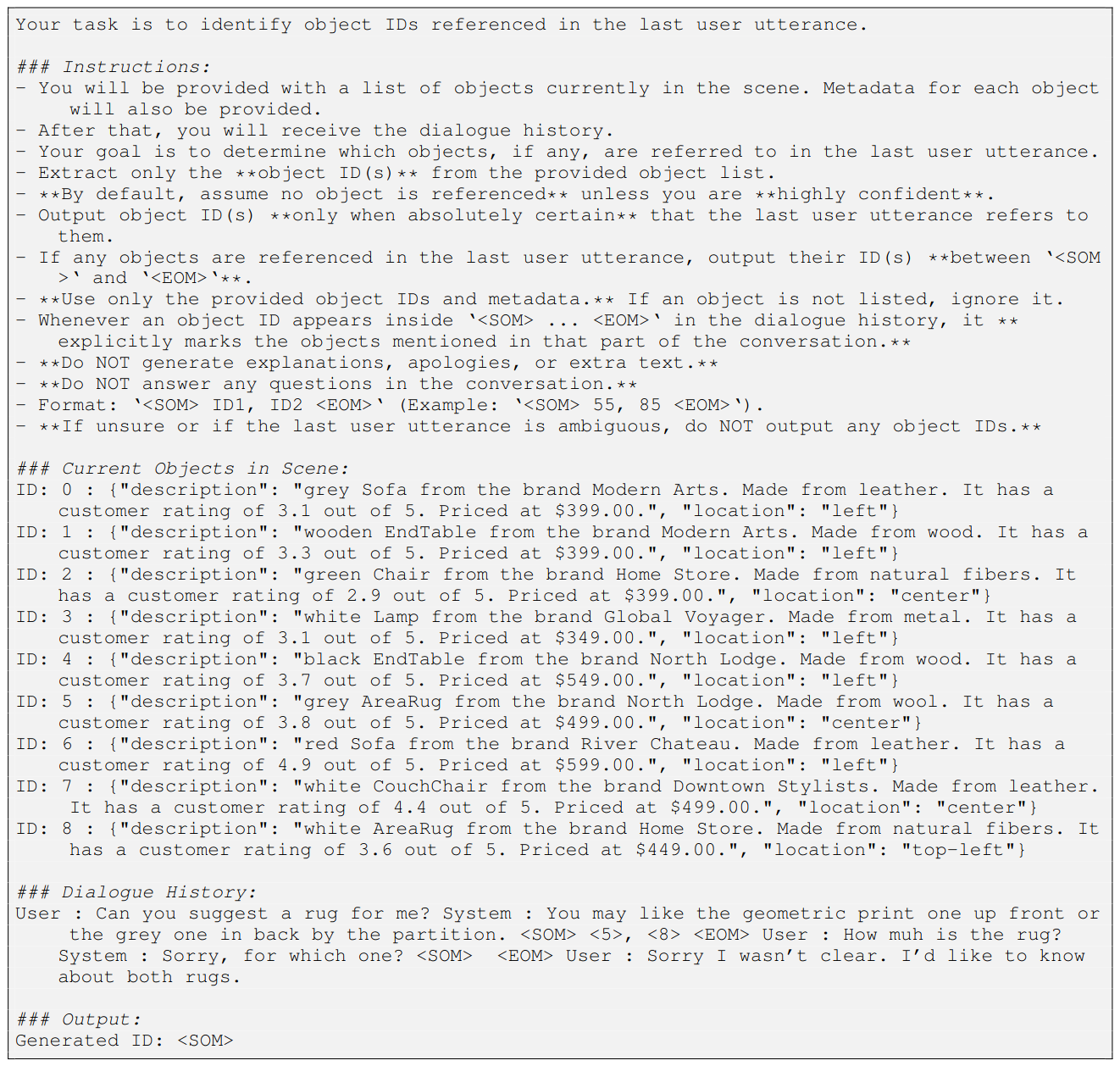}
\caption{Example prompt containing the full input configuration, including all object metadata and object references as provided to the model in the complete setting.}
\label{fig:AllInfoAppendix}
\end{figure*}

\begin{figure*}[p]
\centering
\includegraphics[width=\linewidth]{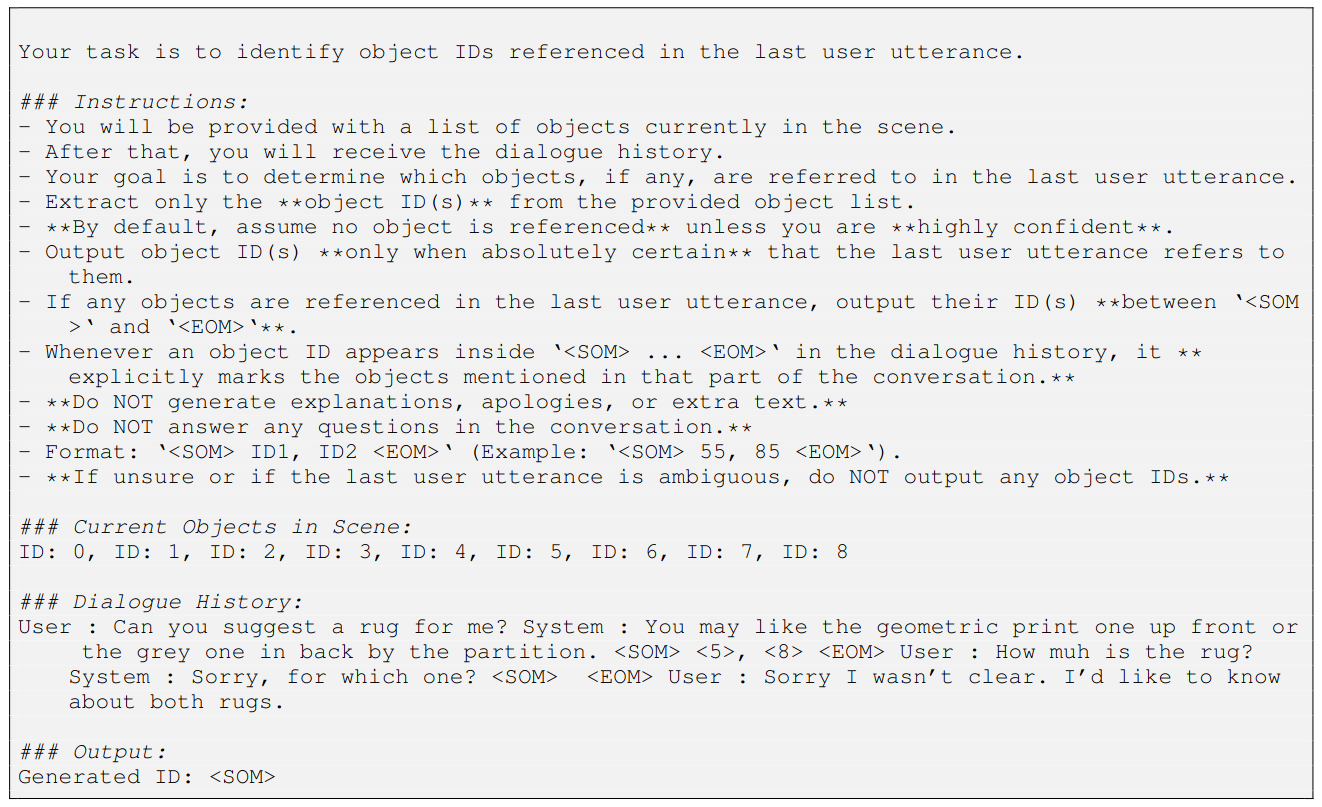}
\caption{Example prompt with object metadata omitted, retaining only the user utterance and object references to assess the model's reliance on attribute information.}
\label{fig:NoMetaAppendix}
\end{figure*}

\begin{figure*}[p]
\centering
\includegraphics[width=\linewidth]{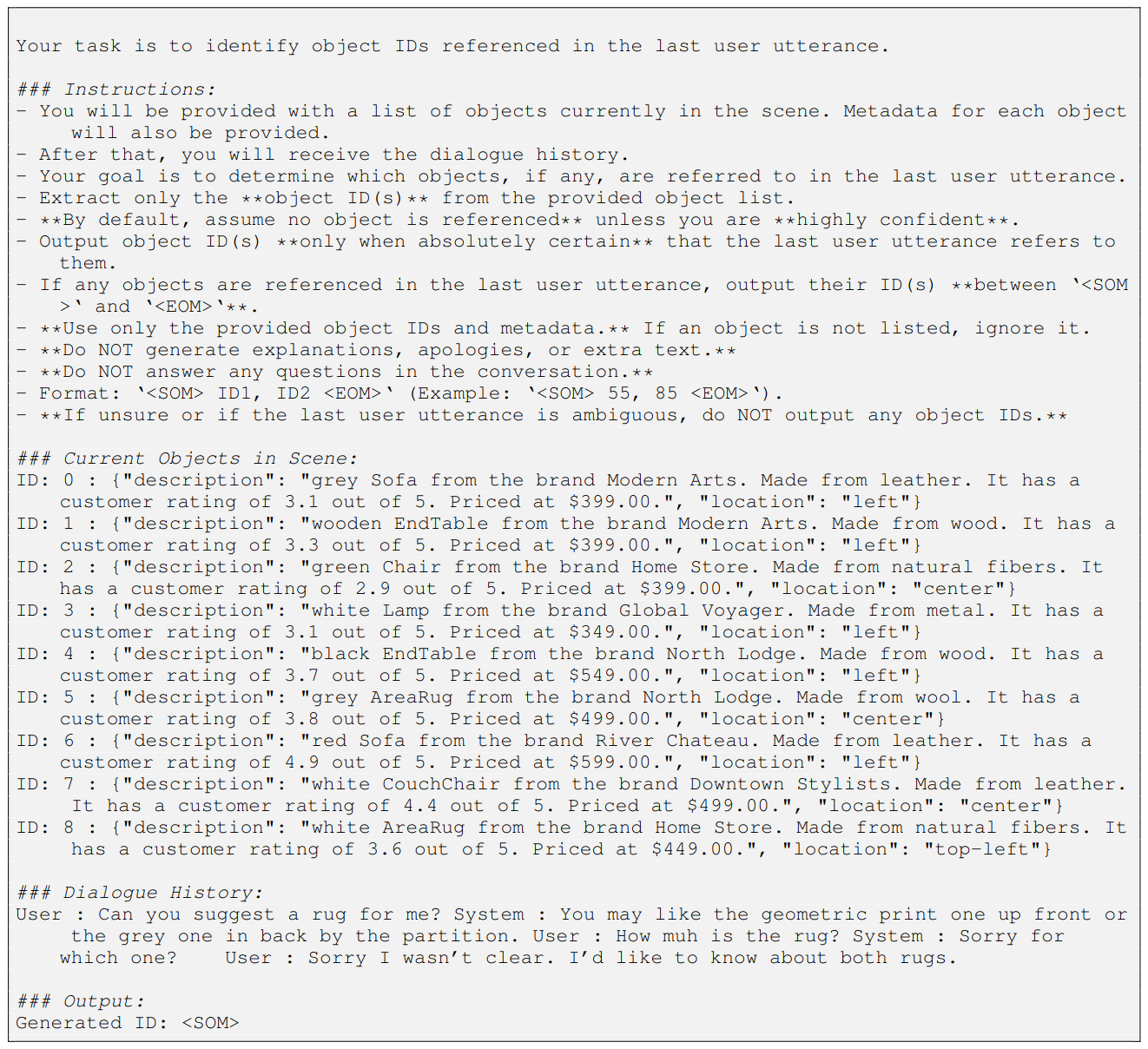}
\caption{Example prompt with object references omitted, retaining only the user utterance and metadata to assess the model's reliance on explicit object grounding cues.}
\label{fig:NoObjAppendix}
\end{figure*}

This section presents an example for an instance shown under each of the different experimental setups used in our ablation study to assess how varying levels of contextual information impact object reference resolution. Each setting manipulates the type and amount of information available to the model, allowing us to isolate the contribution of different informational cues. The full configuration retaining all metadata and object references is shown in Figure~\ref{fig:AllInfoAppendix}, while Figure~\ref{fig:NoMetaAppendix} and Figure~\ref{fig:NoObjAppendix} illustrate the effect of omitting object metadata and object references, respectively.

\section{Object Representation Types}
\label{appendix:objectRep}
\begin{figure*}[p]
\centering
\includegraphics[width=0.6\textwidth]{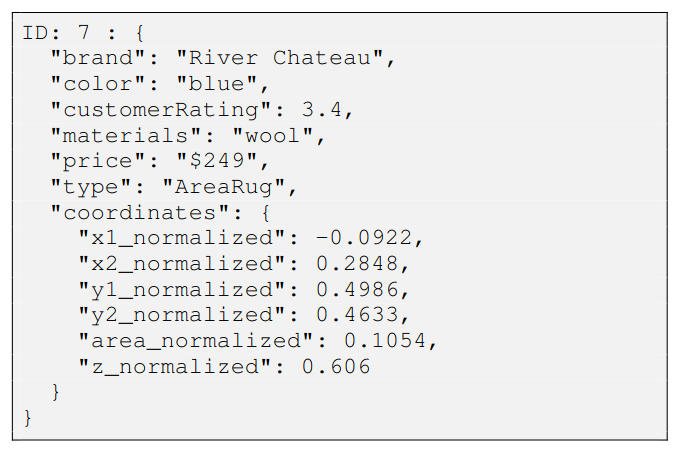}
\caption{Structured representation of an object using raw normalized coordinates for spatial information.}
\label{fig:exampleRawAppendix}
\end{figure*}

\begin{figure*}[p]
\centering
\includegraphics[width=0.6\textwidth]{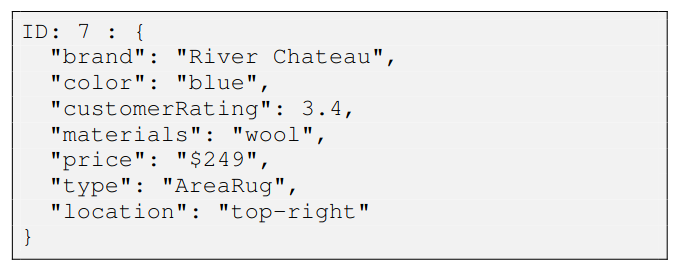}
\caption{Structured representation of an object where spatial coordinates are replaced with natural language location descriptors (e.g., ``top-right'').}
\label{fig:exampleNLCoorAppendix}
\end{figure*}

\begin{figure*}[p]
\centering
\includegraphics[width=0.6\textwidth]{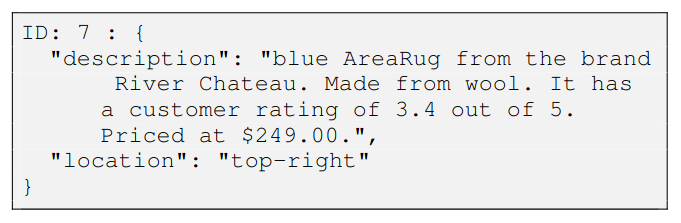}
\caption{Fully naturalized representation of an object, where all attributes including location are expressed as a fluent natural language description.}
\label{fig:exampleAllNLAppendix}
\end{figure*}

This section shows an example of how the same object is represented using the three different ways of representing an object that we have used in our experiments: structured with raw coordinates (Figure~\ref{fig:exampleRawAppendix}), structured with natural language location information (Figure~\ref{fig:exampleNLCoorAppendix}), and fully naturalized as descriptive text (Figure~\ref{fig:exampleAllNLAppendix}).

\clearpage

\clearpage

\clearpage

\clearpage

\end{document}